\documentclass[letterpaper, 10 pt, conference]{ieeeconf} 

\IEEEoverridecommandlockouts                        

\usepackage{graphics}
\usepackage{amsmath} 
\usepackage{amssymb}
\usepackage{cite}
\usepackage{tikz}
\usepackage{amsmath}
\usepackage{tabularx}
\usepackage{array}
\usepackage{booktabs}
\usepackage{bm}
\usepackage{algorithm}
\usepackage{algorithmic}
\usepackage{bbm}
\usepackage{hyperref}

\usepackage{tablefootnote}

\DeclareMathOperator*{\argmax}{argmax}
\DeclareMathOperator*{\argmin}{argmin}
\DeclareMathOperator*{\mode}{mode}

\title{\LARGE \bf
APPLD: Adaptive Planner Parameter Learning from Demonstration
}

\author{Xuesu Xiao$^{1*}$, Bo Liu$^{1*}$, Garrett Warnell$^{2}$, Jonathan Fink$^{2}$, and Peter Stone$^{1}$
\thanks{$^{*}$Equally contributing authors}
\thanks{$^{1}$Xuesu Xiao, Bo Liu, and Peter Stone are with Department of Computer Science, University of Texas at Austin, Austin, TX 78712 {\tt\scriptsize \{xiao, bliu, pstone\}@cs.utexas.edu}}
\thanks{$^{2}$Garrett Warnell and Jonathan Fink are with the Computational and Information Sciences Directorate, Army Research Laboratory, Adelphi, MD 20783 {\tt\scriptsize \{garrett.a.warnell.civ, jonathan.r.fink3.civ\}@mail.mil}}%
}

\begin{document}
\maketitle
\thispagestyle{empty}
\pagestyle{empty}

\begin{abstract}
Existing autonomous robot navigation systems allow robots to move from one point to another in a collision-free manner. However, when facing new environments, these systems generally require re-tuning by expert roboticists with a good understanding of the inner workings of the navigation system. In contrast, even users who are unversed in the details of robot navigation algorithms can generate desirable navigation behavior in new environments via teleoperation. In this paper, we introduce \textsc{appld}, {\em Adaptive Planner Parameter Learning from  Demonstration}, that allows existing navigation systems to be successfully applied to new complex environments, given only a human-teleoperated demonstration of desirable navigation. \textsc{appld} is verified on two robots running different navigation systems in different environments. Experimental results show that \textsc{appld} can outperform navigation systems with the default and expert-tuned parameters, and even the human demonstrator themselves.

\end{abstract}

\section{INTRODUCTION}
\label{sec::introduction}
Designing autonomous robot navigation systems has been a topic of interest to the research community for decades.
Indeed, several widely-used systems have been developed and deployed that allow a robot to move from one point to another \cite{fox1997dynamic, quinlan1993elastic}, often with verifiable guarantees that the robot will not collide with obstacles while moving.

However, while current navigation systems indeed allow robots to autonomously navigate in known environments, they often still require a great deal of tuning before they can be successfully deployed in new environments. 
Adjusting the high-level parameters, or hyper-parameters, of the navigation systems can produce completely different navigation behaviors. 
For example, wide open spaces and densely populated areas may require completely different sets of parameters such as inflation radius, sampling rate, planner optimization coefficients, etc.
Re-tuning these parameters requires an expert who has a good understanding of the inner workings of the navigation system.
Even Zheng's widely-used full-stack navigation tuning guide \cite{zheng2017ros} asserts that fine-tuning such systems is not as simple as it looks for users who are ``sophomoric" about the concepts and reasoning of the system. 
Moreover, tuning a single set of parameters assumes the same set will work well \emph{on average} in different regions of a complex environment, which is often not the case.

In contrast, it is relatively easy for humans---even those with little to no knowledge of navigation systems---to generate desirable navigation behavior in new environments via teleoperation, e.g., by using a steering wheel or joystick.
It is also intuitive for them to adapt their specific navigation strategy to different environmental characteristics, e.g., going fast in straight lines while slowing down for turns.


In this paper, we investigate methods for achieving autonomous robot navigation that are adaptive to complex environments {\em without} the need for a human with expert-level knowledge in robotics.
In particular, we hypothesize that {\em existing} autonomous navigation systems can be successfully applied to complex environments given {\em(1)} access to a human teleoperated demonstration of competent navigation, and {\em(2)} an appropriate high-level control strategy that dynamically adjusts the existing system's parameters.
\begin{figure}[t]
  \centering
  \includegraphics[width=\columnwidth]{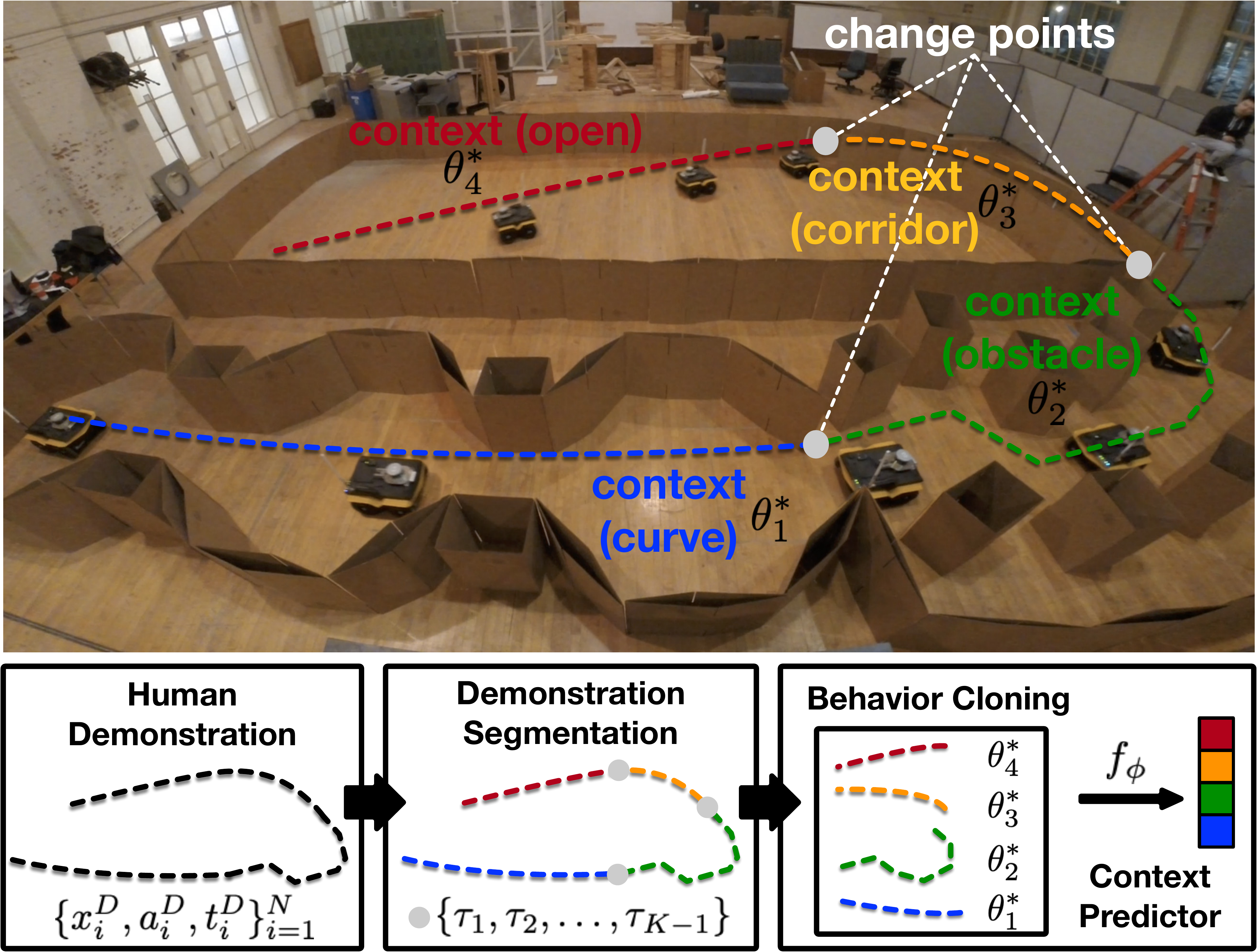}
  \caption{Overview of \textsc{appld}: human demonstration is segmented into different contexts, for each of which, a set of parameters $\theta^*_k$ is learned via Behavior Cloning. During deployment, proper parameters are selected by an online context predictor. Supplementary Video Available at \url{https://www.youtube.com/watch?v=J9AWQWVrjJU}}
  \label{fig:jackal_ahg}
\end{figure}

To this end, we introduce a novel technique called {\em Adaptive Planner Parameter Learning from Demonstration} (\textsc{appld}) and hypothesize that it can outperform default or even expert-tuned navigation systems on multiple robots across a range of environments.
Specifically, we evaluate it on two different robots, each in a different environment, and each using a different underlying navigation system.
Provided with as little as a single teleoperated demonstration of the robot navigating competently in its environment, \textsc{appld} segments the demonstration into contexts based on sensor data and demonstrator behavior and uses machine learning both to find appropriate system parameters for each context and to recognize particular contexts from sensor data alone (Fig. \ref{fig:jackal_ahg}).
During deployment, \textsc{appld} provides a simple control scheme for autonomously recognizing context and dynamically switching the underlying navigation system's parameters accordingly. Experimental results confirm our hypothesis: \textsc{appld} can outperform the underlying system using default parameters and parameters tuned by human experts, and even the performance of the demonstrator.

\section{RELATED WORK}
\label{sec::related_work}

This section summarizes related work on parameter tuning, machine learning for robot navigation, and task demonstration segmentation, also known as changepoint detection.


\subsection{Parameter Tuning}
Broadly speaking, \textsc{appld} seeks to tune the high-level parameters of existing robot navigation systems.
For this task, Zheng's guide \cite{zheng2017ros} describes the current common practice of {\em manual} parameter tuning, which involves robotics experts using intuition, experience, or trial-and-error to arrive at a reasonable set of parameters.
As a result, some researchers have considered the problem of automated parameter tuning for navigation systems, e.g., dynamically finding trajectory optimization weights \cite{teso2019predictive} for a Dynamic Window Approach (\textsc{dwa}) planner \cite{fox1997dynamic}, optimizing two different sets of \textsc{dwa} parameters for straight-line and U-turn scenarios \cite{binch2020context}, or designing novel systems that can leverage gradient descent to match expert demonstrations \cite{bhardwaj2019differentiable}.
While such approaches do successfully perform automatic navigation tuning, they are thus far tightly coupled to the specific system or scenario for which they are designed and typically require hand-engineered features.
In contrast, the proposed automatic parameter tuning work is more broadly applicable: \textsc{appld} treats the navigation system as a black box, and it does not require hand-engineering of features.


\subsection{Machine Learning for Navigation}
Researchers have also considered using machine learning, especially Learning from Demonstration \cite{argall2009survey} or Imitation Learning \cite{hussein2017imitation}, more generally in robot navigation, i.e., beyond tuning the parameters of existing systems.
One such approach is that of using inverse reinforcement learning to estimate costs over driving styles \cite{kuderer2015learning}, social awareness \cite{shiarlis2017rapidly, perez2018teaching, perez2018learning}, and semantic terrain labels \cite{wigness2018robot} from human demonstrations, which can then be used to drive classical planning systems.
Other work has taken a more end-to-end approach, performing navigation by learning functions that map directly from sensory inputs to robot actions \cite{pfeiffer2017perception,siva2019robot}.
In particular, recent work in this space from Kahn {\em et al.} \cite{kahn2020badgr} used a neural network to directly assign costs to sampled action sequences using camera images.
Because these types of approaches seek to replace more classical approaches to navigation, they also forgo the robustness, reliability, and generality of those systems.
For example, Kahn {\em et al.} reported the possibility of catastrophic failure (e.g., flipping over) during training.
In contrast, the work we present here builds {\em upon} traditional robot navigation approaches and uses machine learning to improve them only through parameter tuning, which preserves critical system properties such as safety.



\subsection{Temporal Segmentation of Demonstrations}
\textsc{appld} leverages potentially lengthy human demonstrations of robotic navigation.
In order to effectively process such demonstrations, it is necessary to first segment these demonstrations into smaller, cohesive components.
This problem is referred to as {\em changepoint detection} \cite{aminikhanghahi2017survey}, and several researchers concerned with processing task demonstrations have proposed their own solutions \cite{niekum2015online,fearnhead2007line, meier2012movement, krishnan2017transition, iqbal2019fast}.
Our work leverages these solutions in the context of learning from human demonstrations of navigation behavior.
Moreover, unlike \cite{meier2012movement}, we use the discovered segments to then train a robot for---and deploy it in---a target environment.


\section{APPROACH}
\label{sec::approach}
To improve upon existing navigation systems, the problem considered here is that of \emph{determining a parameter-selection strategy that allows a robot to move quickly and smoothly to its goal}.

We approach this problem as one of learning from human demonstration.
Namely, we assume that a human can provide a teleoperated demonstration of desirable navigation behavior in the deployment environment and we seek to find a set of planner parameters that can provide a good approximation of this behavior.
As we will show in Section \ref{sec::experiments}, when faced with a complex environment, a human demonstrator naturally drives differently in each regions of the environment such that no single set of planner parameters can closely approximate the demonstration in all states.
To overcome this problem, the human demonstration is divided into pieces that include consistent sensory observations and navigation commands. 
By segmenting the demonstration in this way, each piece---which we call a \emph{context}---corresponds to a relatively cohesive navigation behavior.
Therefore, it becomes more feasible to find a single set of planner parameters that imitates the demonstration well for each context.

\subsection{Problem Definition}
We assume we are given a robot with an existing navigation planner $G: \mathcal{X} \times \Theta \rightarrow \mathcal{A}$.
Here, $\mathcal{X}$ is the state space of the planner (e.g., current robot position, sensory inputs, navigation goal, etc.), $\Theta$ is the space of free parameters for $G$ (e.g., sampling density, maximum velocity, etc.), and $\mathcal{A}$ is the planner's action space (e.g., linear and angular velocity). 
Using $G$ and a particular set of parameters $\theta$, the robot performs navigation by repeatedly estimating its state $x$ and applying action $a = G(x;\theta)= G_{\theta}(x)$.
Importantly, we treat $G$ as a black-box, e.g., we do not assume that it is differentiable, and we need not even understand what each component of $\theta$ does.
In addition, a human demonstration of successful navigation is recorded as time series data $\mathcal{D} = \{ x^D_i, a^D_i, t^D_i \}_{i=1}^N$, where $N$ is the length of the series, and $x^D_i$ and $a^D_i$ represent the robot state and demonstrated action at time $t^D_i$.
Given $G$ and $\mathcal{D}$, the particular problem we consider is that of finding two functions: \textbf{(1)} a mapping $M: \mathcal{C} \rightarrow \Theta$ that determines planner parameters for a given {\em context} $c$, and \textbf{(2)} a parameterized context predictor $B_\phi: \mathcal{X} \rightarrow \mathcal{C}$ that predicts the context given the current state.
Given $M$ and $B_\phi$, our system then performs navigation by selecting actions according to $G(x; M(B_\phi(x)))$.
Note that since the formulation presented here involves only changing the {\em parameters} of $G$, the learned navigation strategy will still possess the benefits that many existing navigation systems can provide, such as assurance of safety.

\subsection{Demonstration Segmentation}
Provided with a demonstration, the first step of the proposed approach is to segment the demonstration into pieces---each of which corresponds to a single context only---so that further learning can be applied for each specific context.
This general segmentation problem can be, in principle, solved by any changepoint detection method~\cite{aminikhanghahi2017survey}.
Given $\mathcal{D}$, a changepoint detection algorithm $A_{\text{segment}}$ is applied to automatically detect how many changepoints exist and where those changepoints are within the demonstration.
Denote the number of changepoints found by $A_{\text{segment}}$ as $K-1$ and the changepoints as $\tau_1, \tau_2, \dots, \tau_{K-1}$ with $\tau_0 = 0$ and $\tau_K = N+1$, the demonstration $\mathcal{D}$ is then segmented into $K$ pieces $\{\mathcal{D}_k=\{x^D_i, a^D_i, t^D_i \,|\, \tau_{k-1} \leq i < \tau_k\}\}_{k=1}^K$.

\subsection{Parameter Learning} Following demonstration segmentation, we then seek to learn a suitable set of parameters $\theta^*_k$ for each segment $\mathcal{D}_k = \{x^D_i, a^D_i, t^D_i \,|\, \tau_{k-1} \leq i < \tau_k\}$.
To find this $\theta^*_k$, we employ {\em behavioral cloning} (\textsc{BC}) \cite{pomerleau1989alvinn}, i.e., we seek to minimize the difference between the demonstrated actions and the actions that $G_{\theta_k}$ would produce on $\{x^D_i\}$.
More specifically,
\begin{equation}
    \begin{split}
    \theta^*_k &= \argmin_{\theta} \sum_{(x, a) \in \mathcal{D}_k} ||a - G_{\theta}(x))||_H,
    \end{split}
    \label{eqn:bc}
\end{equation}
where $||v||_H = v^THv$ is the induced norm by a diagonal matrix $H$ with positive real entries, which is used for weighting each dimension of the action. A black-box optimization method $A_{\text{black-box}}$ is then applied to solve Equation \ref{eqn:bc}. Having found each $\theta^*_k$, the mapping $M$ is simply $M(k) = \theta^*_k$.

\subsection{Online Context Prediction}
At this point, we have a library of learned parameters and the mapping $M$ that is responsible for mapping a specific context to its corresponding parameters. All that remains is a scheme to dynamically infer which context the robot is in during \emph{execution}. To do so, we form a supervised dataset $\{x^D_i, c_i\}_{i=1}^N$, where $c_i = k$ if $x^D_i \in \mathcal{D}_k$. Then, a parameterized function $f_\phi(x)$ is learned via supervised learning to classify which segment $x^D_i$ comes from, i.e., 
\begin{equation}
    \phi^* = \argmax_\phi \sum_{i=1}^N \log \frac{\exp\big(f_\phi(x^D_i)[c_i]\big)}{\sum_{c=1}^K \exp{\big(f_\phi(x^D_i)[c]\big)}}.
\end{equation}
Given $f_\phi$, we define our context predictor $B$ according to
\begin{equation}
    B_\phi(x_t) = \mode \Big\{\argmax_{c} f_\phi(x_i)[c],\,\, t-p < i \leq t \Big\}.
    \label{eqn:mode_switching}
\end{equation}
In other words, $B_\phi$ acts as a mode filter on the context predicted by $f_\phi$ over a sliding window of length $p$.

\begin{algorithm}[t]
\small
\caption{\textsc{appld}} \label{alg:appld}
\begin{algorithmic}[1]
    \STATE{// Training}
    \STATE{\textbf{Input:} the demonstration $\mathcal{D} = \{ x^D_i, a^D_i, t^D_i \}_{i=1}^N$, space of possible parameters $\Theta$, and the navigation stack $G$}.
    \STATE Call $A_{\text{segment}}$ on $\mathcal{D}$ to detect changepoints $\tau_1, \dots, \tau_{K-1}$ with $\tau_0=0$ and $\tau_K=N+1$.
    \STATE Segment $D$ into $\{\mathcal{D}_k=\{x^D_i, a^D_i, t^D_i \,|\, \tau_{k-1} \leq i < \tau_k\}\}_{k=1}^K$.
    \STATE Train a classifier $f_\phi$ on $\{x^D_i, c_i\}_{i=1}^N$, where $c_i = k$ if $x^D_i \in \mathcal{D}_k$.
    \FOR{$k=1:K$}
        \STATE Call $A_{\text{black-box}}$ with objective defined in Equation \ref{eqn:bc} on $\mathcal{D}_k$ to find parameters $\theta^*_k$ for context $k$.
    \ENDFOR
    \STATE Form the map $M(k) = \theta^*_k$, $\forall 1\leq k\leq K$.
    \STATE 
    \STATE{// Deployment}
    \STATE \textbf{Input:} the navigation stack $G$, the mapping $M$ from context to parameters, and the context predictor $B_\phi$.
    \FOR{$t=1:T$}
        \STATE Identify the context $c_t = B_\phi(x_t)$ according to Equation \ref{eqn:mode_switching}.
        \STATE Navigate with $G(x_t; M(c_t))$.
    \ENDFOR
\end{algorithmic}
\end{algorithm}

Taken together, the above steps constitute our proposed \textsc{appld} approach. During training, the above three stages are applied sequentially to learn a library of parameters $\{\theta^*_k\}_{k=1}^K$ (hence the mapping $M$) and a context predictor $B_\phi$. During execution, Equation \ref{eqn:mode_switching} is applied online to pick the right set of parameters for navigation. Algorithm \ref{alg:appld} summarizes the entire pipeline from offline training to online execution.

\section{EXPERIMENTS}
\label{sec::experiments}
\begin{figure*}[t]
  \centering
  \includegraphics[width=\textwidth]{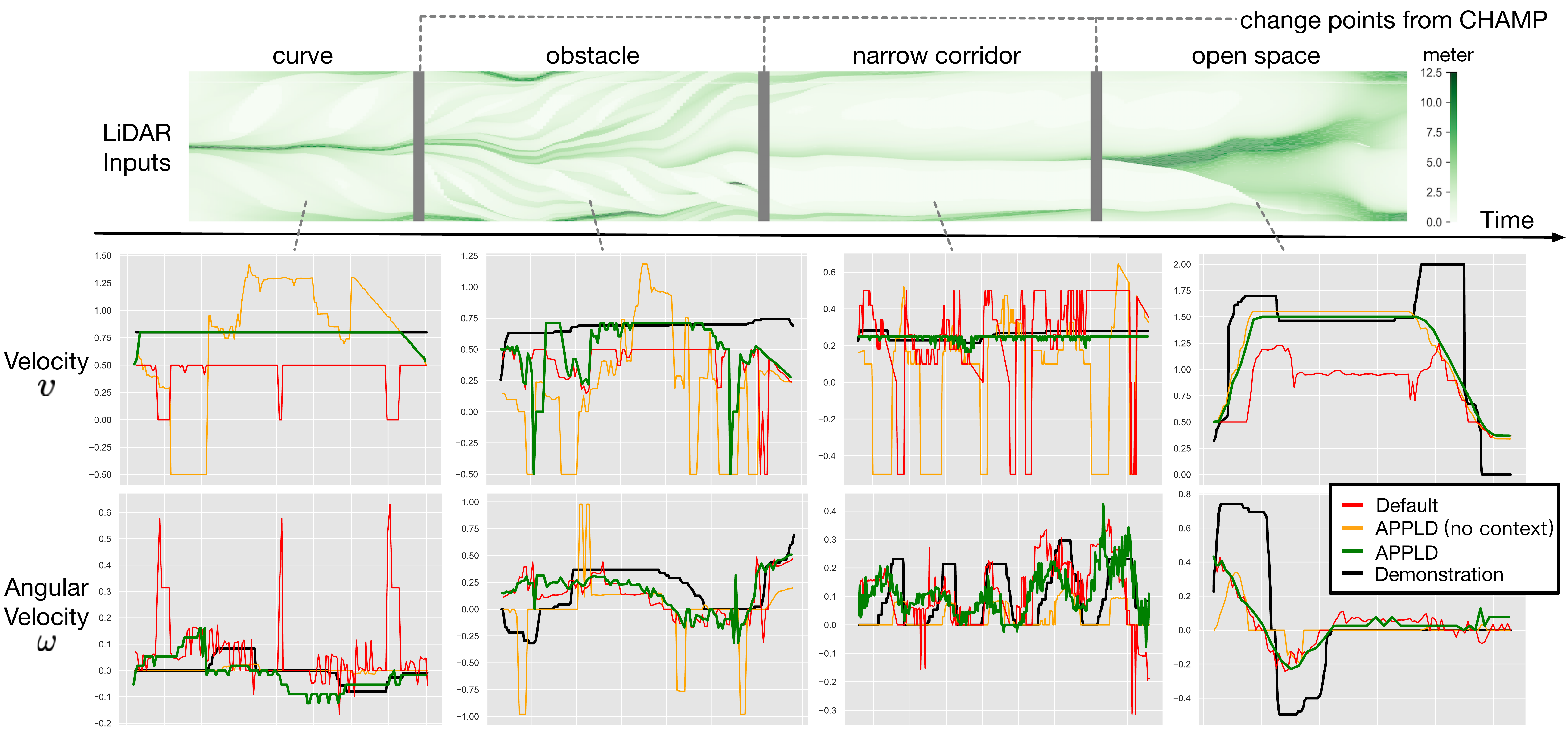}
  \caption{Jackal Trajectory in Environment Shown in Fig. \ref{fig:jackal_ahg}: heatmap visualization of the LiDAR inputs over time is displayed at the top and used for segmentation by \textsc{champ}. For each region divided by \textsc{champ} changepoints, CMA-ES finds a set of parameters that best imitates the human demonstration. Velocity and angular velocity profiles from \textsc{default} (red), \textsc{appld (no context)} (orange), and \textsc{appld} (green) parameters, along with the human demonstration (black), are displayed with respect to time. Plots are scaled to best demonstrate performance differences between different parameters.}
  \label{fig:ahg4modes}
\end{figure*}

In this section, \textsc{appld} is implemented to experimentally validate our hypothesis that {\em existing} autonomous navigation systems can be successfully applied to complex environments given {\em (1)} access to a human demonstration from teleoperation, and {\em (2)} an appropriate high-level control strategy that dynamically adjusts the existing system's parameters based on context.
To perform this validation, \textsc{appld} is applied on two different robots---a Jackal and a BWIBot---that each operate in a different environment with different underlying navigation methods.
The results of \textsc{appld} are compared with those obtained by the underlying navigation system using {\em (a)} its default parameters (\textsc{default}) from the robot platform manufacturer, and {\em (b)} parameters we found using behavior cloning but without context (\textsc{appld (no context)}).
We also compare to the navigation system as tuned by robotics experts in the second experiment. In all cases, we find that \textsc{appld} outperforms the alternatives.

\subsection{Jackal Maze Navigation}
In the first experiment, a four-wheeled, differential-drive, unmanned ground vehicle---specifically a Clearpath Jackal---is tasked to move through a custom-built maze (Fig. \ref{fig:jackal_ahg}).
The Jackal is a small and agile platform with a top speed of 2.0m/s.
To leverage this agility, the complex maze consists of four qualitatively different areas: {\em (i)} a pathway with curvy walls (curve), {\em (ii)} an obstacle field (obstacle), {\em (iii)} a narrow corridor (corridor), and {\em (iv)} an open space (open) (Fig. \ref{fig:jackal_ahg}).
A Velodyne LiDAR provides 3D point cloud data, which is transformed into 2D laser scan for 2D navigation. 
The Jackal runs Robot Operating System (\textsc{ros}) onboard, and \textsc{appld} is applied to the local planner, \textsc{dwa} \cite{fox1997dynamic}, in the commonly-used \texttt{move\textunderscore base} navigation stack. Other parts of the navigation stack, e.g. global planning with Dijkstra's algorithm, remain intact.

Teleoperation commands are captured via an Xbox controller from one of the authors with previous experience with video games, who is unfamiliar with the inner workings of the \textsc{dwa} planner and attempts to operate the platform to quickly traverse the maze in a safe manner. The teleoperator follows the robot and controls the robot from a third person view.
This viewpoint, different from the robot's first person view, may provide the human demonstrator with different contextual information, but our experiments will show that the robot's limited onboard LiDAR input suffices for online context identification. 
The 52s demonstration is recorded using \texttt{rosbag} configured to record all joystick commands and all inputs to the \texttt{move\textunderscore base} node. 

For changepoint detection (Algorithm \ref{alg:appld}, line 3), we use \textsc{champ} as $A_{\text{segment}}$, a state-of-the-art Bayesian segmentation algorithm \cite{niekum2015online}.
The recorded LiDAR range data statistics (mean and standard deviation) from $X_i^D$ and the recorded demonstrated actions $a_i^D=(v_i^D, w_i^D)$ are provided as input to \textsc{champ}.
\textsc{champ} outputs a sequence of changepoints $\tau_1, \tau_2, \dots, \tau_{K-1}$ that segment the demonstration into $K$ segments, each with uniform context (line 4).
As expected, \textsc{champ} determines $K=4$ segments in the demonstration, each corresponding to a different context (line 5).
$f_{\phi}$ trained for online context prediction (line 14) is modeled as a two-layer neural network with ReLU activation functions.

\begin{figure}[]
  \centering
  \includegraphics[width=1\columnwidth]{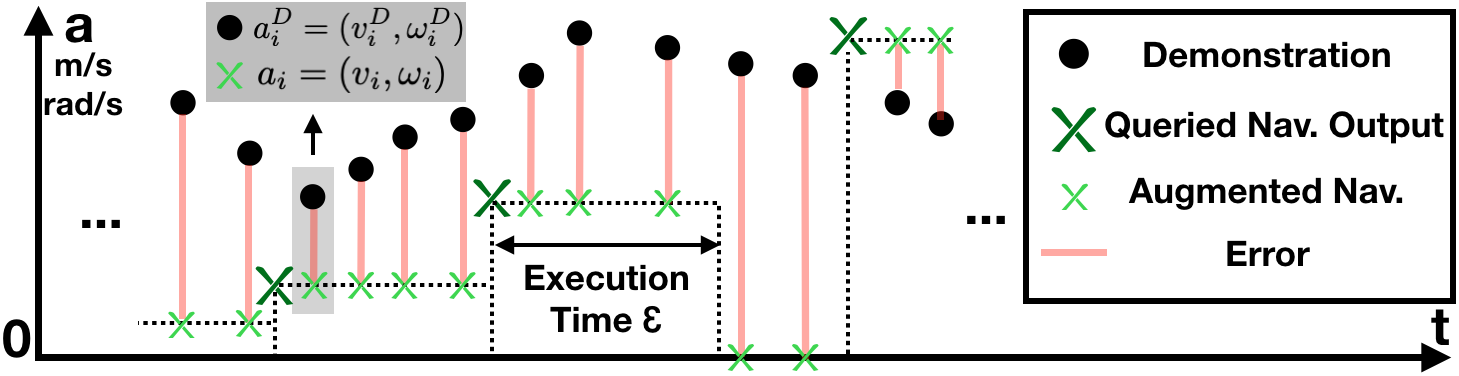}
  \caption{Action-Matching and Loss Metric}
  \label{fig:mse}
\end{figure}

For the purpose of finding $\theta_k^*$ for each context, the recorded input is played to a \textsc{ros} \texttt{move\textunderscore base} node using \textsc{dwa} as the local planner with query parameters $\theta$ and the resulting output navigation commands are compared to the demonstrator's actions.
Ideally, the \textsc{dwa} output and the demonstrator commands would be aligned in time, but for practical reasons (e.g., computational delay), this is generally not the case---the output frequency of \texttt{move\textunderscore base} is much lower than the frequency of recorded joystick commands.
To address this discrepancy, we match each $a^D_i$ with the {\em most recent} queried output of $G_{\theta}$ within the past $\epsilon$ seconds (default execution time per command, 0.25s for Jackal), and use it as the augmented navigation at time $t^D_i$.
If no such output exists, augmented navigation is set to zero since the default behavior of Jackal is to halt if no command has been received in the past $\epsilon$ seconds (Fig. \ref{fig:mse}).
This condition may occur due to insufficient onboard computation to perform sampling at the requested density.
For the metric in Equation \ref{eqn:bc}, we use mean-squared error, i.e. $H$ is the identity matrix.

Following the action-matching procedure, we find each $\theta_k^*$ using \textsc{cma-es} \cite{hansen2003reducing} as our black-box optimizer (Algorithm \ref{alg:appld}, line 7).
The optimization runs on a single Dell XPS laptop (Intel Core i9-9980HK) using 16 parallel threads.
The elements of $\theta$ in our experiments are: \textsc{dwa}'s \emph{max\_vel\_x} (v), \emph{max\_vel\_theta} (w), \emph{vx\_samples} (s), \emph{vtheta\_samples} (t), \emph{occdist\_scale} (o), \emph{pdist\_scale} (p), \emph{gdist\_scale} (g) and costmap's \emph{inflation\_radius} (i).
We intentionally select parameters here that directly impact navigation behavior and exclude parameters which are robot-model-specific, e.g. physical acceleration limit (\emph{acc\_lim\_x} and  \emph{acc\_lim\_theta}), or unrelated to the behaviors being studied, e.g. goal tolerance (\emph{xy\_goal\_tolerance}). Note that \emph{max\_vel\_x} and \emph{max\_vel\_theta} are not the physical velocity limit of the robot, but rather the maximum velocity commands that are allowed to be executed. They interact with the sampling density parameters, \emph{vx\_samples} and \emph{vtheta\_samples}, in a way that affects whether finding a reasonable motion command through sampling can be performed in real time. The parameters \emph{occdist\_scale}, \emph{pdist\_scale}, and \emph{gdist\_scale}, are optimization weights for distance to obstacle, distance to path, and distance to goal, respectively. The inflation radius, \emph{inflation\_radius}, specifies the physical safety margin to be used around obstacles.
All parameters are initialized at the midpoint between their lower- and upper-bound. The fully parallelizable optimization takes approximately eight hours, but this time could be significantly reduced with more computational resources and engineering effort.

The action profiles of using the parameters discovered by \textsc{default}, \textsc{appld (no context)}, and \textsc{appld} are plotted in Fig. \ref{fig:ahg4modes}, along with the single-shot demonstration segmented into four chunks by \textsc{champ}. 
Being trained separately based on the segments discovered by \textsc{champ}, the \textsc{appld} parameters (green) perform most closely to the human demonstration (black), whereas the performance of both \textsc{default} (red) and \textsc{appld (no context)} (orange) significantly differs from the demonstration in most cases (Fig. \ref{fig:ahg4modes}).

\begin{table}[h]
  \caption{Parameters of Jackal Experiments (\textsc{dwa}):  \newline \emph{max\_vel\_x } \textnormal{(v)}, \emph{max\_vel\_theta} \textnormal{(w)}, \emph{vx\_samples} \textnormal{(s)}, \emph{vtheta\_samples} \textnormal{(t)}, \emph{occdist\_scale} \textnormal{(o)}, \emph{pdist\_scale} \textnormal{(p)}, \emph{gdist\_scale \textnormal{(g)}}, \emph{inflation\_radius \textnormal{(i)}}}
  \label{tab:jackal_parameters}
  \centering
  \small
  \begin{tabular}{lrrrrrrrr}
    \toprule
                & v & w & s & t & o & p & g & i \\
    \midrule
    \textsc{def.}       & 0.50 & 1.57 &  6 & 20 & 0.10 & 0.75 & 1.00 & 0.30 \\
    \textsc{no ctx.} & 1.55 & 0.98 & 10 &  3 & 0.01 & 0.87 & 0.99 & 0.46 \\
    \midrule
    Curve       & 0.80 & 0.73 &  6 & 42 & 0.04 & 0.98 & 0.94 & 0.19 \\
    Obstacle    & 0.71 & 0.91 & 16 & 53 & 0.55 & 0.54 & 0.91 & 0.39 \\
    Corridor    & 0.25 & 1.34 &  8 & 59 & 0.43 & 0.65 & 0.98 & 0.40 \\
    Open  & 1.59 & 0.89 & 18 & 18 & 0.40 & 0.46 & 0.27 & 0.42 \\
    \bottomrule
  \end{tabular}
\end{table}

The specific parameter values learned by each technique are given in Tab. \ref{tab:jackal_parameters}, where we show in the bottom rows the individual parameters learned by \textsc{appld} for each context. 
The learned parameters relative to the default values are intuitive in many ways. For example, \textsc{appld} found that Curve requires a larger value for the parameters p and g and a lower value for the parameter i, i.e., the platform needs to place a high priority on sticking to the straight global path so that it can avoid extraneous motion due to the proximity of the curvy walls. It is similarly intuitive that \textsc{appld} found that Obstacle Field requires higher sampling rates (s and t) and more consideration given to obstacle avoidance (higher o) in order to find feasible motion through the irregular obstacle course. Corridor is extremely tight, and, accordingly, \textsc{appld} found that a smaller linear velocity (v) was necessary in order to compensate for the larger computational load associated with the necessary higher angular velocity sampling rate (t) required to find feasible paths. In Open, \textsc{appld} appropriately learned that the maximum velocity (v) should be increased in order to match the demonstrator's behavior. In addition to these intuitive properties, \textsc{appld} was also able to capture other, more subtle, parameter interactions that are more difficult to describe.
At run time, \textsc{appld}'s trained context classifier selects in which mode the navigation stack is to operate and adjusts the parameters accordingly (Fig. \ref{fig:jackal_ahg}).

Tab. \ref{tab:agh4modes} shows the results of evaluating the overall navigation system using the different parameter-selection schemes along with the demonstrator's performance as a point of reference.
We report both the time it takes for each system navigate a pre-specified route and also the \textsc{BC} loss (Equation \ref{eqn:bc}) compared to the demonstrator. We choose to study traversal time since most suboptimal navigation behavior will cause stop-and-go motions, induce recovery behaviors, cause the robot to get stuck, or collide with obstacles (termination) -- each of which will result in a higher traversal time.
\begin{table}[h]
  \caption{Loss and Time Comparison of Jackal Experiments (\textsc{dwa})}
  \label{tab:agh4modes}
  \centering
  \small
  \begin{tabular}{lrr}
    \toprule
    Context        & BC Loss                             & Real-world Time (s)\\
    \midrule
    (a) Curve\\
    Demonstration           & N/A                                 & 12.10 \\
    \textsc{default}                       & 0.1755$\pm$0.0212                   & 30.20$\pm$3.87 \\
    \textsc{app. (no ctx.)}   & 0.1856$\pm$0.0030                   &  *55.14$\pm$13.84 \\
    \textsc{appld}         & \textbf{0.0780}$\pm$\textbf{0.0002} & \textbf{9.39}$\pm$\textbf{0.73}\\
    \midrule
    (b) Obstacle Field\\
    Demonstration           & N/A                                 & 9.00 \\
    \textsc{default}                       & 0.2061$\pm$0.0540                   & 12.32$\pm$0.59 \\
    \textsc{app. (no ctx.)}   & 0.2537$\pm$0.0083                   & *60.00$\pm$0.00 \\
    \textsc{appld}         & \textbf{0.1586}$\pm$\textbf{0.0216} & \textbf{7.69}$\pm$\textbf{0.35} \\
    \midrule
    (c) Narrow Corridor\\
    Demonstration           & N/A                                 & 24.06 \\
    \textsc{default}                       & 0.0953$\pm$0.0916                   & *49.52$\pm$19.88 \\
    \textsc{app. (no ctx.)}   & 0.0566$\pm$0.0419                   & *60.00$\pm$0.00 \\
    \textsc{appld}         & \textbf{0.0198}$\pm$\textbf{0.0010} & \textbf{19.11}$\pm$\textbf{0.82} \\
    \midrule
    (d) Open Space\\
    Demonstration           & N/A                                 & 7.28 \\
    \textsc{default}                       & 0.8597$\pm$0.0013                   & 15.07$\pm$0.61 \\
    \textsc{app. (no ctx.)}   & 0.2094$\pm$0.0013                   & 15.08$\pm$7.42 \\
    \textsc{appld}         & \textbf{0.2071}$\pm$\textbf{0.0021} & \textbf{7.19}$\pm$\textbf{0.42} \\
    \bottomrule
  \end{tabular}
\end{table}

For each metric, lower is better, and we compute mean and standard deviation over 10 independent trials.
For trials that end in failure (e.g., the robot gets stuck), we add an asterisk (*) to the reported results and use penalty time value of 60s.
The results show that, for every context, \textsc{appld} achieves the lowest \textsc{BC} loss and fastest real-world traverse time, compared to \textsc{default} and \textsc{appld (no context)}.
In fact, while \textsc{appld} is able to successfully navigate in every trial, \textsc{default} fails in 8/10 trials in the narrow corridor due to collisions in \texttt{recovery\textunderscore behaviors} after getting stuck, and \textsc{appld (no context)} fails in 9/10, 10/10, and 10/10 trials in curve, obstacle field, and narrow corridor, respectively.
In open space, \textsc{appld (no context)} is able to navigate quickly at first, but is not able to precisely and quickly reach the goal due to low angular sample density (\emph{vtheta\_samples}). Surprisingly, in all contexts, the navigation stack with \textsc{appld} parameters even outperforms the human demonstration in terms of time, and leads to qualitatively smoother motion than the demonstration. Average overall traversal time from start to goal, 43s, is also faster than the demonstrated 52s. 
The superior performance achieved by \textsc{appld} compared to \textsc{default} and even the demonstrator validates our hypothesis that given access to a teleoperated demonstration, tuning \textsc{dwa} navigation parameters is possible without a roboticist. We notice that, in some challenging situations, even the human demonstrator suffered from suboptimal navigation, e.g. stop-and-go, overshoot, etc. Even in these cases, \textsc{appld} can produce smooth, stable, and sometimes even faster navigation due to the benefit of a properly-parameterized autonomous planner. The fact that \textsc{appld} outperforms \textsc{appld (no context)} indicates the necessity of the high-level context switch strategy.

\subsection{BWIBot Hallway Navigation}
\begin{figure}[t]
  \centering
  \includegraphics[width=1\columnwidth]{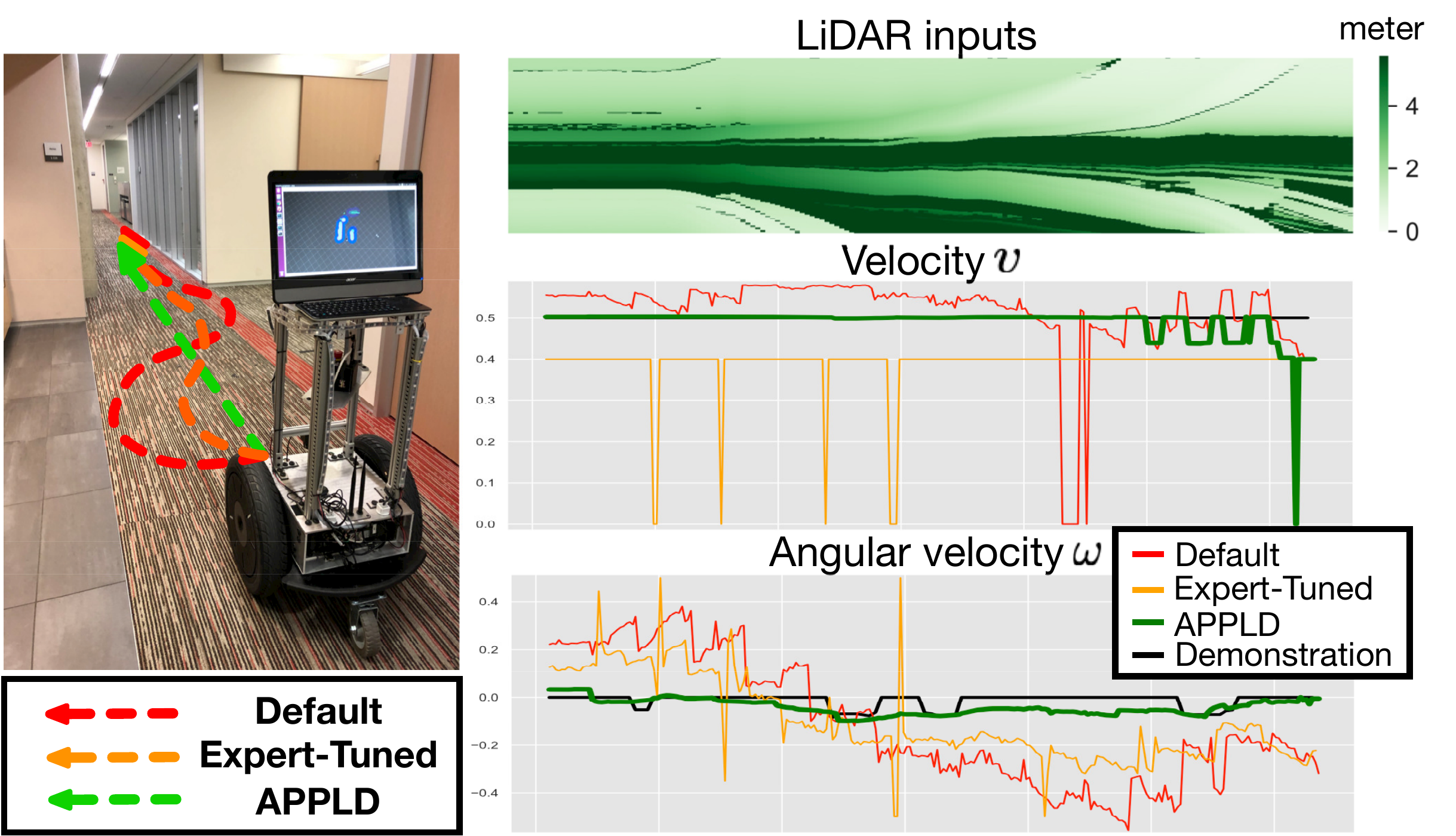}
  \caption{BWIBot Navigates in GDC Hallway}
  \label{fig:gdc}
\end{figure}
Whereas we designed the Jackal experiments to specifically test all aspects of \textsc{appld}, in this section, we evaluate \textsc{appld}'s generality to another robot in another environment running another underlying navigation system.
Specifically, we evaluate our approach using a BWIBot (Fig. \ref{fig:gdc} left)---a custom-built robot that navigates the GDC building at The University of Texas at Austin every day as part of the Building Wide Intelligence (BWI) project \cite{khandelwal2017bwibots}.
The BWIBot is a nonholonomic platform built on top of a Segway RMP mobile base, and is equipped with a Hokuyo LiDAR.
A Dell Inspiron computer performs all computation onboard.
Similar to the Jackal, the BWIBot uses the \textsc{ros} architecture and the \texttt{move\textunderscore base} navigation framework.
However, unlike the Jackal, the BWIBot uses a local planner based on the elastic bands technique (\textsc{e-band}) \cite{quinlan1993elastic} instead of \textsc{dwa}.

As in the Jackal experiments, teleoperation is performed using an Xbox controller from a third person view by the same author who is unfamiliar with the inner workings of the \textsc{e-band} planner. 
The demonstration lasts 17s and consists of navigating the robot through a hallway, where the demonstrator seeks to move the robot in smooth, straight lines at a speed appropriate for an office environment. Unlike the Jackal demonstration, quick traversal is not the goal of the demonstration. 

In this setting, the \textsc{appld} training procedure is identical to that described for the Jackal experiments.
In this case, however, \textsc{champ} did not detect any changepoints based on the LiDAR inputs and demonstration (Fig. \ref{fig:gdc} right), indicating the hallway environment is relatively uniform and hence one set of parameters is sufficient.

The \textsc{BC} phase takes about two hours with 16 threads on the same laptop used for the Jackal experiments.
The parameters learned for the \textsc{e-band} planner are \emph{max\_vel\_lin} (v), \emph{max\_vel\_th} (w), \emph{eband\_internal\_force\_gain} (i), \emph{eband\_external\_force\_gain} (e), and \emph{costmap\_weight} (c). The results are shown in Tab. \ref{tab:bwi_results}.

\begin{table}[t]
  \caption{Parameters and Results of BWIBot Experiments (\textsc{e-band}): \newline \emph{max\_vel\_lin} \textnormal{(v)}, \emph{max\_vel\_th} \textnormal{(w)}, \emph{eband\_internal\_force\_gain} \textnormal{(i)}, \emph{eband\_external\_force\_gain} \textnormal{(e)}, \emph{costmap\_weight} \textnormal{(c)}}
  \label{tab:bwi_results}
  \centering
  \small
  \begin{tabular}{lrrrrrr}
    \toprule
                  & v & w & i & e & c & loss\\
    \midrule
    \textsc{def.}     & 0.75 & 1.0  &    1 &    2 & 10    & 0.1730$\pm$0.0025\\
    \textsc{exp.}     & 0.5  & 0.5  &    3 &    2 & 15    & 0.0940$\pm$0.0095\\
    \midrule
    \textsc{app.}     & 0.65 & 0.35 & 0.52 & 0.04 & 15.36 & \textbf{0.0669}$\pm$\textbf{0.0071}\\
    \bottomrule
  \end{tabular}
\end{table}

The first row of Tab. \ref{tab:bwi_results} shows the parameters of the BWIBot planner used in the \textsc{default} system.
Because \textsc{champ} does not discover more than a single context, \textsc{appld} and \textsc{appld (no context)} are equivalent for this experiment.
Therefore, we instead compare to a set of expert-tuned (\textsc{expert}) parameters that is used on the robot during everyday deployment, shown in the second row of the table. 
These parameters took a group of roboticists several days to tune by trial-and-error to make the robot navigate in relatively straight lines. 
Finally, the parameters discovered by \textsc{appld} are shown in the third row. 
The last column of the table shows the \textsc{BC} loss induced by \textsc{default}, \textsc{expert}, and \textsc{appld} parameters (again averaged over 10 runs).
Real-world time is not reported since a quick traversal is not the purpose of the demonstration in the indoor office space. 
The action profiles from these three sets of parameters (queried on the demonstration trajectory $\{ x^D_i \}_{i=1}^N$) are compared with the demonstration and plotted in Fig. \ref{fig:gdc} lower right, where the learned trajectories are the closest to the demonstration.
When tested on the real robot, the \textsc{appld} parameters achieve qualitatively superior performance, despite the fact that the experts were also trying to make the robot navigate in a straight line (Fig. \ref{fig:gdc} left).


The BWIBot experiments further validate our hypothesis that parameter tuning for existing navigation systems is possible based on a teleoperated demonstration instead of expert roboticist effort. 
More importantly, the success on the \textsc{e-band} planner without any modifications from the methodology developed for \textsc{dwa} supports \textsc{appld}’s generality.

\section{SUMMARY AND FUTURE WORK}
\label{sec::conclusions}

This paper presents \textsc{appld}, a novel learning from demonstration framework that can autonomously learn suitable planner parameters and adaptively switch them during execution in complex environments.
The first contribution of this work is to grant non-roboticists the ability to tune navigation parameters in new environments by simply providing a single teleoperated demonstration.
Secondly, this work allows mobile robots to utilize existing navigation systems, but adapt them to different contexts in complex environments by adjusting their navigation parameters on the fly.
\textsc{appld} is validated on two robots in different environments with different navigation algorithms.
We observe superior performance of \textsc{appld}'s parameters compared with all tested alternatives, both on the Jackal and the BWIBot. 
An interesting direction for future work is to investigate methods for speeding up learning by clustering similar contexts together. It may also be possible to perform parameter learning and changepoint detection jointly for better performance.

\section*{ACKNOWLEDGEMENT}
This work has taken place in the Learning Agents Research Group (LARG) at UT Austin. LARG research is supported in part by NSF (CPS-1739964, IIS-1724157, NRI-1925082), ONR (N00014-18-2243), FLI (RFP2-000), ARL, DARPA, Lockheed Martin, GM, and Bosch.  Peter Stone serves as the Executive Director of Sony AI America and receives financial compensation for this work.  The terms of this arrangement have been reviewed and approved by the University of Texas at Austin in accordance with its policy on objectivity in research.

\bibliographystyle{IEEEtran}
\bibliography{IEEEabrv,references}
\end{document}